\documentclass{article}

    \PassOptionsToPackage{numbers, compress}{natbib}

\usepackage[final]{neurips_2025}

\renewcommand{\cite}[1]{\citep{#1}}

\usepackage[utf8]{inputenc} 
\usepackage[T1]{fontenc}    
\usepackage{hyperref}       
\usepackage{url}            
\usepackage{booktabs}       
\usepackage{amsfonts}       
\usepackage{nicefrac}       
\usepackage{microtype}      
\usepackage{xcolor}         
\usepackage{natbib}
\usepackage{makecell}
\usepackage{subcaption}
\usepackage{authblk}
\usepackage{siunitx}
\usepackage{colortbl}
\newcommand{\gray}[1]{{\cellcolor{lightgray}} #1}

\title{msf-CNN: Patch-based Multi-Stage Fusion with  Convolutional Neural Networks for TinyML}

\author[1]{Zhaolan~Huang \thanks{Corresponding Author. E-mail: zhaolan.huang@fu-berlin.de} }
\author[1, 2, 3]{Emmanuel~Baccelli}

\affil[1]{\small Freie Universität Berlin}
\affil[2]{\small Einstein Center Digital Future}
\affil[3]{\small Inria, France}

\usepackage{graphicx}
\usepackage[final]{changes}

\usepackage{xspace}

\usepackage{amsmath}
\usepackage{amssymb}
\usepackage{mathtools}
\usepackage{amsthm}

\usepackage[capitalize,noabbrev]{cleveref}

\theoremstyle{plain}

\theoremstyle{definition}

\theoremstyle{remark}

\newcommand{\methodName}{msf-CNN\xspace}

\usepackage{multirow}
\usepackage{acronym}
\newacro{AI}{Artificial Intelligence}
\newacro{DAG}{directed acyclic graph}
\newacro{FL}{Federated Learning}
\newacro{ML}{Machine Learning}
\newacro{NN}{Neural Network}
\newacro{CNN}{Convolutional Neural Network}
\newacro{MPU}{memory protection unit}
\newacro{TMLaaS}{TinyML-as-a-Service}
\newacro{IoT}{Internet of Things}
\newacro{CBOR}{Concise Binary Object Representation}
\newacro{CoAP}{Constrained Application Protocol}
\newacro{IID} {independent and identically distributed} 
\newacro{MCU}{Microcontroller Unit}
\newacro{I$^2$C}{Inter-Integrated Circuit}
\newacro{IETF}{Internet Engineering Task Force}
\newacro{MLP}{Multi-Layer Perceptron} 
\newacro{ANN}{Artifical Neural Networks}
\newacro{TinyML}{Tiny Machine Learning}
\newacro{AIoT}{Artificial Intelligence of Things}
\newacro{ROC}{Receiver Operating Characteristic}
\newacro{AUC}{Area Under the Curve}
\newacro{WASN}{Wireless acoustic sensor network}
\newacro{ASN}{Acoustic sensor network}
\newacro{PTQ}{Post-Training Quantization}
\newacro{TPR}{True Positive Rate}
\newacro{FPR}{False Positive Rate}
\newacro{STFT}{Short-Time Fourier Transform}
\newacro{SNR}{Signal-to-noise ratio}
\newacro{DC}{Direct Current}
\newacro{OOM}{out-of-memory}
\newacro{NAS}{Neural Architecture Search}
\newacro{TVM}{Tensor Virtual Machine}
\newacro{DNN}{Deep Neural Network}
\newacro{MAC}{multiply–accumulate}
\newacro{RAM}{Random Access Memory}
\newacro{IR}{intermediate representation}
\newacro{SOTA}{state-of-the-art}
\newacro{DRAM}{Dynamic RAM}
\newacro{FPGA}{Field Programmable Gate Array}

\begin{document}

\framebox[1.01\width]{\parbox{\dimexpr\linewidth-2\fboxsep-2\fboxrule}{If you cite this paper, please use the NeurIPS 2025 reference: Z. Huang and E. Baccelli. msf-CNN: Patch-based Multi-Stage Fusion with  Convolutional Neural Networks for TinyML. in Proceedings of the 39th Conference on Neural Information Processing Systems (NeurIPS 2025).}}

\maketitle

\begin{abstract}

AI spans from large language models to tiny models running on microcontrollers (MCUs).
Extremely memory-efficient model architectures are decisive to fit within an MCU's tiny memory budget e.g., 128kB of RAM.
However, inference latency must remain small to fit real-time constraints.
An approach to tackle this is \emph{patch-based fusion}, which aims to optimize data flows across neural network layers.
In this paper, we introduce \emph{\methodName}, a novel technique that efficiently finds optimal fusion settings for convolutional neural networks (CNNs) by walking through the fusion solution space represented as a directed acyclic graph.
Compared to previous work on CNN fusion for MCUs, \methodName identifies a wider set of solutions.
We published an implementation of \methodName running on various microcontrollers (ARM Cortex-M, RISC-V, ESP32). We show that \methodName can achieve inference using 50\% less RAM compared to the prior art (MCUNetV2 and StreamNet). We thus demonstrate how \methodName offers additional flexibility for system designers.

\end{abstract}

\section{Introduction}

\ac{AIoT} is a domain aiming to embed AI in the smallest networked devices~\cite{ghosh2018artificial}.
As such \ac{AIoT} is pushing the miniaturization of \acp{DNN} to fit microcontroller-based hardware, which enables various applications at the edge of the network. Use-cases include vision/audio recognition, environmental monitoring, personalized medical care, etc. However, imbalance between the increasing resource requirements of \acp{DNN} and the very limited computation capacity (CPU in MHz) and memory resource of \acp{MCU} remains a challenge in deploying \acp{DNN} on \ac{IoT} devices. For instance, as described in RFC7228~\cite{rfc7228}, billions of \ac{IoT} devices are resource-constrained devices, with \ac{RAM} smaller than 50 KiB, and Flash memory smaller than 250 KiB. On the other hand, even a single convolutional layer in quantized ResNet-34~\cite{koonce2021resnet-34, he2016deepresnet} consumes around 414.72 KiB in \ac{RAM}. This example highlights the huge gap between memory budgets on \ac{IoT} devices and \ac{RAM} usage of \acp{DNN}.

A technique aimed at decreasing this gap is patch-based layer \emph{fusion}, introduced in~\cite{alwani_fused-layer_2016}. Initially targeting FPGAs, patch-based fusion reduces off-chip \ac{DRAM} requirements and communication bus transfer costs for inference with CNNs. Fusion is great for low-memory devices because it can save up to 95\% of \ac{RAM} usage. Moreover, Fusion decouples input size from memory usage, allowing for larger input. Recent work has thus explored the use of fusion on \acp{MCU}, for example, to improve the memory consumption of the first few convolutional layers of MobileNetV2~\cite{lin_memory-efficient_2021}.

Nevertheless, we observe that significant issues linger on \acp{MCU}. First, intermediate feature maps inside the fusing block incur a high (re)compute cost. Second, input size limits hamper many use-cases such as medical image processing, sequence time series analysis (e.g. audio application), etc. Third, implementations of fusion on \acp{MCU} have so far been very hardware-specific (e.g. bound to the ARM-Cortex-M7 instruction set) and model-specific (e.g. bound to CNN mobile inverted blocks).

{\bf Contributions -- }
With the goal of improving on the above issues, we report on following work:

\begin{itemize}
\itemsep 0em 

\item We propose \methodName, a fusion-based approach to achieve ultra-low \ac{RAM} footprint of neural network inference and we open-source its implementation\footnote{Please check \url{https://github.com/TinyPART/msf-CNN}};

\item We formulate the problem of finding optimal fusion settings that minimize peak \ac{RAM} usage or compute cost of neural networks as a variant shortest path problem.

\item We provide graph models representing multi-stage fusion neural networks, which encode peak \ac{RAM} usage and compute cost of single and fused layers.
\item We designed a pruning strategy to squeeze the search space and use graph-based algorithm to find solutions in reasonable time complexity (from $O(2^{N-2})$ to \replaced{polynomial time}{$O(N^2)$}).
\item We improved global pooling and dense operators to further squeeze \ac{RAM} usage without compute overhead.

\item We released preliminary evaluation results on MCU-based IoT boards. We compared common CNN, StreamNet, MCUNetV2 and \methodName on a variety of microcontrollers. We show that \methodName allows new trade-off between memory saving and compute overhead.


\end{itemize}


While our main focus is on microcontrollers, msf-CNN is not limited to them. Its cost estimator and C-code backend also support general CPU platforms (e.g., x86, Cortex-A), enabling broader use such as memory-optimized cloud inference. Moreover, with appropriate cost models and backends, msf-CNN can be extended to accelerators (GPUs, FPGAs, ASICs). We leave non-MCU experiments to future work to maintain focus on devices with tighter RAM constraints.

\section{Background}
\label{sec:background}

\textbf{Patch-based Fusion for DNN on FPGA \& GPU -- }
Patch-based fusion was initially proposed in \cite{alwani_fused-layer_2016} as a fusion scheme for \ac{CNN} deployed on \ac{FPGA} to reduce the off-chip \ac{DRAM} usage and I/O overhead. Instead of computing the complete feature maps for each layer, it fuses convolutional layers into a single block (pyramid structure) and computes only one or a few output elements. This approach requires only small portions (tiles) of the feature maps loaded onto \ac{DRAM}. However, the reduction of \ac{RAM} is at the cost of re-computing the overlapped elements in feature maps required by adjacent fused layers. DeFiNES~\cite{mei2023defines}, another fusion framework, explored different cache strategies within fused layers to alleviate the re-computation issue. (Fully-recompute, H-Cached \& V-recompute, and Fully-cache).
Fully-recompute eliminates caching entirely, requiring all overlapping input tensor elements to be recalculated; H-cached \& V-recompute caches elements along the horizontal axis while recomputing vertical overlaps; and Fully-cache retains all overlapping elements in memory. These approaches illustrate a critical trade-off—enhanced caching progressively reduces compute redundancy but proportionally increases \ac{RAM} usage, with cached element quantity inversely correlating to compute overhead and directly scaling with memory demands. Additional work has also applied fusion on GPUs, for instance~\cite{pinckaers_streaming_2022} used it for cancer detection in medical pictures.
\added{Note that \emph{patch-based} fusion is fundamentally different from \emph{kernel} fusion techniques. We elaborate on this distinction and its implications in \cref{appdix:rel_msf_kernel_fusion}.}
    

\textbf{Patch-based Fusion on MCUs -- } 
Work on MCUNetV2 \cite{lin_memory-efficient_2021} has applied fusion on MobileNetV2 to reduce the peak \ac{RAM} usage. It revealed that layers at the head of the model dominate the \ac{RAM} usage. Hence these layers were fused into one block to reduce \ac{RAM} usage significantly. The recompute issue was mitigated by redistributing the receptive field, so the receptive field inside the fusion block was decreased and regained at a later stage. Work on StreamNet \cite{zheng2024streamnet} introduced a two-dimensional tensor cache to significantly reduce re-compute operations in a fusion block and applied brute force to search for optimal fusion position and cache depth. Nevertheless, no prior work explored the potential of multiple fusion blocks in CNNs.

\textbf{Representing DNNs as Inverted Dataflow Graphs -- }
\label{subsec:nn_as_dag}
Dataflow graph have been widely used for modeling \ac{DNN}, as pioneered by TensorFlow and PyTorch \cite{abadi2016tensorflow, paszke2019pytorch}. The data (tensor) flows alongside the directed edge between nodes which indicates the operations (convolution, pooling, addition, etc.) applied on the incoming edges (tensors). This representation shows the producer-consumer relations among operations and has great expressiveness and flexibility, enabling automatic differentiation and concurrent execution of independent operations.

\section{High-Level Idea}
Inspired by the above previous works, \methodName aims to answer the following questions: \textbf{(1)} Where to fuse and how to determine the fusion position/depth? \textbf{(2)} Under specific resource constraints, how to find the optimal fusion settings?

\begin{figure}[htbp]
  \centering
  \begin{subcaptionbox}{Overview\label{fig:overview}}[0.34\linewidth]
    {\includegraphics[width=\linewidth]{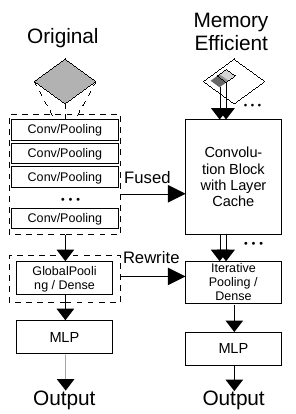}}
  \end{subcaptionbox}
  \hfill
  \begin{subcaptionbox}{Neural Network as DAG\label{fig:nn_as_dag}}[0.64\linewidth]
    {\includegraphics[width=\linewidth]{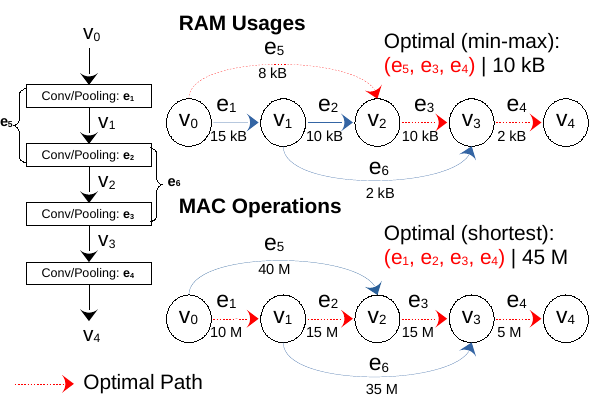}}
  \end{subcaptionbox}
  \caption{\textbf{(a)} Overview of msf-CNN. The convolutional layers are fused into several fusion blocks based on the optimal setting found by optimizer. We let global Pooling and dense layers compute the outputs iteratively to further squeeze \ac{RAM} usage. \textbf{(b)} The neural network is modeled as a \ac{DAG}. Nodes $v_n$ denote the tensors that are produced and consumed by the operators or possible fusion blocks. Edges $e_1, \dots, e_4$ represent individual operators, while edge $e_5, e_6$ represent two candidate fusion blocks. Edges are annotated with the \ac{RAM} usage and \ac{MAC} amounts of their corresponding operators and fusion blocks.}
  \label{fig:main}
\end{figure}

As depicted in \cref{fig:overview}, \methodName determines fusion settings (fusion position and depth), transforms layers accordingly into fusion blocks and rewrites global pooling and dense layers as their iterative implementation, which can further squeeze \ac{RAM} usage without any computation overhead.

To guide us in doing so, we use \textit{inverted} dataflow graphs to model CNNs, where tensors are represented as nodes, and operations are depicted as edges connecting them. On this graph, we encode into the edges the resource usage of the operations, and use additional edges to represent fusion blocks. This allows us to design graph-based strategies to find optimal solutions with lower computational complexity using proven graph algorithms.


\section{Problem Definitions \& Assumptions}
\label{subsec:prb_statement}

We aim to solve a pair of dual optimization problems. Let $\chi$ be the set of all possible configurations for fusion blocks. We \textbf{define \textit{P1} as the problem of minimizing peak \ac{RAM} usage} subject to a computation cost limit:
\begin{align} \label{equ:min_peak_mem}
    & \min_{S} P(\chi, S) \\
    & \text{s.t. } F(\chi, S) < F_{max} \label{equ:mac_limit}
\end{align}
where $P$ is the peak \ac{RAM} usage, and $F$ is the computation overhead for inference under fusion setting $S$, relatively to inference without fusion (thereafter denoted \emph{vanilla}). The compute cost limit and \ac{RAM} limit are annotated by $F_{max}$ and $P_{max}$, respectively. Dually, we \textbf{define \textit{P2} as the problem of minimizing computation cost} subject to a \ac{RAM} footprint limit:
\begin{align} \label{equ:min_compute_cost}
    & \min_{S} F(\chi, S) \\
    & \text{s.t. } P(\chi, S) < P_{max} \label{equ:ram_limit}
\end{align}

Without loss of generality, we only discuss fusion blocks of convolutions. We assume a \textit{H-Cache} scheme, which we chose to be a good trade-off between buffer size and recompute cost on \acp{MCU}. 

In \cref{appdix:cache_buffer_size} and \cref{appdix:mac_compute}, we further detail the analysis of the cache buffer size and the amount of \ac{MAC} operations.

\section{DNN Graph Representation \& Formulation}
We interpret the optimization problems described in \cref{subsec:prb_statement} by modeling the \ac{DNN} as data-nodes graph. We transform the problem as a shortest path problem \cite{sedgewik2001graph_algorithms} and use off-the-shelf graph algorithms to find a solution that minimizes the peak memory usage as well as compute cost during inference regarding specified constraints.

\subsection{DNN Representation}
As described in \cref{subsec:nn_as_dag}, we model a \ac{DNN} as a \ac{DAG} $G=(V, E)$ with data nodes $v_0, \dots, v_n$ representing input/output tensors of consecutive layers and $m$ edges $e_1, \dots, e_m$ that represent single layers or fusion blocks. Each edge is also encoded with resource requirements by layer or fusion block. Specifically, the first ($v_0$) and the last node ($v_n$) are the input and output tensor of the neural network, respectively. 

In general, the edge represents the input/output relation of nodes and also indicates the fusion depth inside the neural network. For example, an edge that connects consecutive vertices $e=v_{n} \rightarrow v_{n+1}$ is a single layer that consumes $v_{n}$ as input tensor and outputs tensor $v_{n+1}$, while an edge that jumps over multiple vertices $e=v_{n} \rightarrow v_{n+m}, m > 1$ represents a fusion block with $m$ layers. Each \textbf{complete compute path} from $v_0$ to $v_n$ represents a fusion setting $S$.

A typical example depicted in \cref{fig:nn_as_dag} explains how to use \ac{DAG} for representing a simple neural network. Tensors are transformed into nodes, operators and fusion blocks are edges. Edges are encoded with \ac{RAM} usages and \ac{MAC} amounts of their corresponding operators. Hence, the problem is transformed to find an optimal path from the input node to the output node of the graph.


\subsection{Encoding RAM Usage}
We first calculate the \ac{RAM} usages $P_{e_i}$ of all single layers and all possible fusion blocks inside the neural network by
\begin{equation}
    P_{e_i} = I + O + Buf
\end{equation}
where $I$ and $O$ are the size of input and output tensor, respectively. $Buf$ represents the cache buffer size of the fusion block, which is determined by the chosen cache scheme. In this work it is given in \cref{equ:buf_size_ith_layer}. Trivially, for non-fused layers $Buf$ is always set to zero since no fusion cache is needed.

Thereafter, the calculated \ac{RAM} usages are attached to the corresponding edges for further analysis. For a complete compute path contains $n$ edges $S=(e_{i_1}, \dots, e_{i_n})$ we can then calculate the overall peak \ac{RAM} usage $P_S$ by
\begin{equation}
    P_S = \max_{j=1 \dots n} P_{e_{i_j}}
\end{equation}

\subsection{Encoding Compute Cost}
The encoding steps of compute cost are similar to encoding peak memory usage. Here we use \ac{MAC} operations as the indicator of compute cost. In this paper, the \ac{MAC} amount of fusion block is given in \cref{equ:mac_fused_layer} and \cref{equ:mac_fusion_block}.

After attaching the calculated \acp{MAC} to the edges, the total compute cost of a complete compute path $S$ is
\begin{equation}
    C_S = \sum_{j=1}^{n} C_{e_{i_j}}
\end{equation}
Therefore, the \textbf{compute overhead factor} $F$ representing the ratio of the \ac{MAC} amount after fusion to the vanilla, common one without fusion is expressed as
   $ F = C_S / C_{vanilla} $.
For the constraints in \cref{equ:mac_limit}, users can set a maximum compute overhead factor $F_{max}$ expressed as
   $ F_{max} = C_{max} / C_{vanilla} $.
In the following sections, we will discuss several graph-based algorithms to solve the optimization problem.

\section{Searching for Optimal Fusion Settings}
After building an inverted dataflow graph of a \ac{DNN} with all possible fusion combinations (edges), the two dual problems are indeed transformed into classic graph problems: finding an optimal complete compute path from the input tensor node $v_1$ to the output tensor node $v_n$ under specific constraints.

\textbf{Impact of Search Space Size -- }
If we consider the unconstrained optimization, the solution is trivial: the single-source-single-target shortest path, which can be found by classical graph algorithm like Dijkstra's \cite{dijkstra2022note} with the time complexity of $\mathcal{O}(E \log{(V)})$. However, when considering the constraints, it is necessary first to explore all possible complete compute paths that meet the conditions, which can potentially explode the complexity to $\mathcal{O}(2^{V-2})$ \cite{robert2002algorithms} in the worst case. Hence, we need a smarter strategy to squeeze the search space and avoid horrendous complexity.

\added{Note that such shortest path computations do not take place on the microcontroller at runtime. Instead, they are computed offline, on a PC, which expands the realm of what can be assessed as bearable computation.}

\subsection{Problem P1: Minimizing Peak RAM Usage}
The unconstrained optimization is to find a complete compute path with minimal peak \ac{RAM} usage, which is equivalent to finding the path that minimizes the maximum weight of edges (minimax path problem). As mentioned above, this can be solved by modified Dijkstra algorithm. An example path with minimal peak \ac{RAM} usage is presented in \cref{fig:nn_as_dag}.

For the constraint of compute cost limit (\cref{equ:mac_limit}), the pruning strategy needs co-design with its optimization problem (\cref{equ:min_peak_mem}). We noticed that all possible peak \ac{RAM} usages have already been encoded into the edges. Therefore, the problem can be transformed into the following: we first construct a \textbf{candidate solution set} $\{S_0, S_1, \dots, S_i, \dots \}$ with
\begin{align}
    & S_i = \arg\min_S C(G_i, S), \\
    & G_i := \text{subgraph of $G_{i-1}$, obtained by removing} \notag \\
    & \text{all edges in $G_{i-1}$ with the maximal RAM usage}, \\
    & G_0 = G
\end{align}
where $C(G_i, S)$ is the \ac{MAC} amount of fusion setting $S$ in graph $G_i$. The candidate solution $S_i$ can be obtained by applying the shortest path algorithm. We then filter the candidate solutions to find those that satisfy the constraints and select the one with the smallest \ac{RAM} usage as the optimal solution.

In this way, we avoid constructing a search space with a complexity of 
$\mathcal{O}(2^{V-2})$. Instead, we iteratively eliminate subgraphs and solve for candidate solutions, reducing the complexity to $\mathcal{O}(V^3)$. For most deep neural networks running on \acp{MCU}, this process can be done in few seconds.

\subsection{Problem P2: Minimizing Compute Cost}
We first discuss the unconstrained variant, which is identical to $P_{max} = \infty$. In this case, finding the solution is equivalent to finding the shortest complete compute path -- the path with a minimal sum of \ac{MAC} -- of the graph, which can be again solved by classical algorithm like Dijkstra's \cite{dijkstra2022note}. \cref{fig:nn_as_dag} shows an example with an optimal path marked in red.

When bringing back the constraint of \ac{RAM} limit, the pruning step is simple: eliminating all edges with encoded \ac{RAM} usage exceeding the limit. So, all paths in the graph will automatically fulfill the limitation.

\subsection{Analytical Results}
\label{sec:analy_rslt}
To explore the capability of these two dual optimizers, here we choose three variants of MobileNetV2 and MCUNet \cite{sandler_mobilenetv2_2018, lin_memory-efficient_2021} with different scales for the pilot study: MobileNetV2 with width multiplier 0.35 and input size of $144 \times 144 \times 3$ (MBV2-w0.35), MCUNetV2-VVW-5fps with input size of $80 \times 80 \times 3$ (MN2-vvw5), MCUNetV2-320KB-ImageNet with input size of  $176 \times 176 \times 3$ (MN2-320K). For optimizer of minimizing peak \ac{RAM} usage, the maximal compute overhead factor ranges from 1.1 to 1.5 then jumps to Infinite, which represents an unconstrained optimization. For optimizer of minimizing compute cost, the maximal peak \ac{RAM} usage was set from 16 kB to 256 kB where each level represents a popular \ac{RAM} capacity of mainstream \acp{MCU}.

As shown in \cref{tbl:anly_rslt}, both optimizers can indeed theoretically suppress the peak \ac{RAM} usage without violating all preset constraints. The high \ac{RAM} usage compression is achieved with increase of deep fusion blocks, thereby introducing a high compute overhead. The extreme cases lay on the unconstrained optimization minimizing the \ac{RAM} usage by more than $90 \%$, while reluctantly introducing $1.6 \times$ to $2.7 \times$ of compute overhead. This is only suitable for time-intensive applications with a high limited \ac{RAM} budget.

On the other hand, setting appropriate constraints can still lead to well-optimized configurations, with our tools offering flexibility to accommodate real-life scenarios. Under different thresholds on compute overhead factor or peak \ac{RAM} usage, the solutions that optimizer found are all fulfill the constraints and with \ac{RAM} usage all lower than the vanilla, un-fused setting. In some cases, it is even possible to compress RAM usage without incurring additional computational overhead. These pilot studies demonstrate the effectiveness of finding usable solutions under real-life constraints.

\added{We also conducted a preliminary analysis of the heuristic strategy used in MCUNetV2, which fuses only the early layers to minimize RAM usage. While this approach is simple and straightforward, it tends to yield suboptimal fusion configurations. As shown in \cref{tbl:anly_rslt}, msf-CNN discovers better solutions and offers greater flexibility compared to the heuristic strategy.} The analytical results were further validated by on-board experiments presented in \cref{sec:experiments}.

\begin{table*}[ht]
\centering
\caption{Analytical results with msf-CNN under different constraints.
Vanilla: un-fused models. \added{Heuristic: minimize RAM consumption by only fusing heading layers.} SAA: Same as above. \colorbox{lightgray}{Gray: msf-CNN beats heuristic.}}
\label{tbl:anly_rslt}
\sisetup{table-format = 2.2, round-mode = places, round-precision = 2}
\begin{tabular}{@{}
                ll
                S
                S
                S
                S
                S
                S
                @{}}
\hline
                      &        & \multicolumn{2}{c}{MBV2-w0.35} & \multicolumn{2}{c}{MN2-vww5} & \multicolumn{2}{c}{MN2-320K}                     \\
                      & Constraint  & \em{RAM (kB)}         & $F$           & \em{RAM (kB)}        & $F$          & \em{RAM (kB)}                  & $F$                    \\\hline
Vanilla &  -   & 194.44           &   1          & 96              &    1        & 309.76                    &   1                   \\
Heuristic &  -   & 32.076              &    1.59     & 24          &   1.56    & 90.307                    &   3.25                 \\\hline
\multirow{6}{*}{P1: $F_{max}$} & 1.1    & 67.905           & 1.1         & 32.792          & 1.04       & 190.096                   & 1.04                 \\
                      & 1.2    & \multicolumn{2}{c}{(SAA)}      & \gray{26.128}          & \gray{1.11}       & 186.736                   & 1.19                 \\
                      & 1.3    & \gray 21.288           & \gray{1.3}         & \gray{17.76}          & \gray{1.3}        & 186.032                   & 1.25                 \\
                      & 1.4    & \gray{15.34}            & \gray{1.38}        & \gray{13.376}          & \gray{1.35}       & 156.672                   & 1.37                 \\
                      & 1.5    & \multicolumn{2}{c}{(SAA)}      & \multicolumn{2}{c}{(SAA)}    & 94.184                    & 1.45                 \\
                      & Inf    & 7.887            & 1.68        & 12              & 1.96       & \gray{42.643}                    & \gray{2.69 }                \\\hline
\multirow{5}{*}{P2: $P_{max}$} & 16 kB  & \gray{15.34}            & \gray{1.38}        & \gray{13.376}          & \gray{1.35}       & \multicolumn{2}{c}{\multirow{2}{*}{(No Solution)}} \\
                      & 32 kB  & 25.674           & 1.25        & \gray{26.128}          & \gray{1.11 }      & \multicolumn{2}{c}{}                             \\
                      & 64 kB  & 63.741           & 1.23        & 38.576          & 1.02       & \gray{62.88}                     & \gray{2.02}                \\
                      & 128 kB & 83.065           & 1.02        & 89.6            & 1          & 94.184                    & 1.45                 \\
                      & 256 kB & 181.44           & 1           & \multicolumn{2}{c}{(SAA)}    & 247.808                   & 1   \\\hline
\end{tabular}

\end{table*}



\section{msf-CNN Implementation Details}



We have implemented the msf-CNN fusion mechanism on top of microTVM v0.16.0~\cite{chen2018tvm}. We use the TVM frontend to convert models into \ac{IR}, and rewrite the compute graph and low-level routines of operators to fit the fusion settings. We used RIOT-ML~\cite{huang2024riot-ml} to benchmark the fused models: the models are transformed into C code using microTVM and integrated in dedicated firmware leveraging a common \ac{IoT} operating system (RIOT~\cite{baccelli2018riot}) which we can run on various boards shown in \cref{tbl:iot_boards}.

\textbf{Sequential RAM Usage -- }
We have optimized the \ac{RAM} usage of the global pooling and fully connected (Dense) layers. We observed that the outputs of these two basic blocks can be computed iteratively, and in most scenarios, their input dimensions are much larger than their output dimensions. As a result, we can temporally divide the input and sequentially process it through the iterative global pooling or dense layers, which further minimizes memory usage. If their upstream is a fusion block, this perfectly matches the feature of temporally split inputs, enabling them to be fused seamlessly.

\begin{figure}[htbp]
  \centering
  \begin{minipage}[b]{0.49\linewidth}
    \centering
    \includegraphics[width=\linewidth]{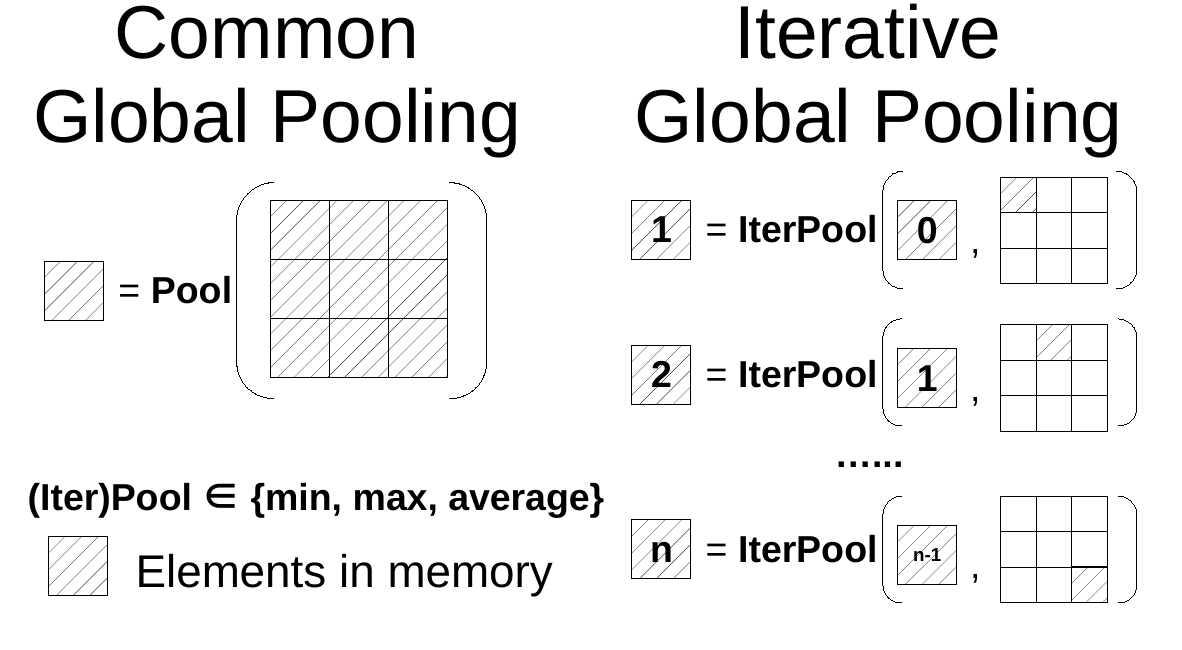}
    \caption{Comparison of common and iterative global pooling.}
    \label{fig:common_vs_iter_pool}
  \end{minipage}
  \hfill
  \begin{minipage}[b]{0.49\linewidth}
    \centering
    \includegraphics[width=\linewidth]{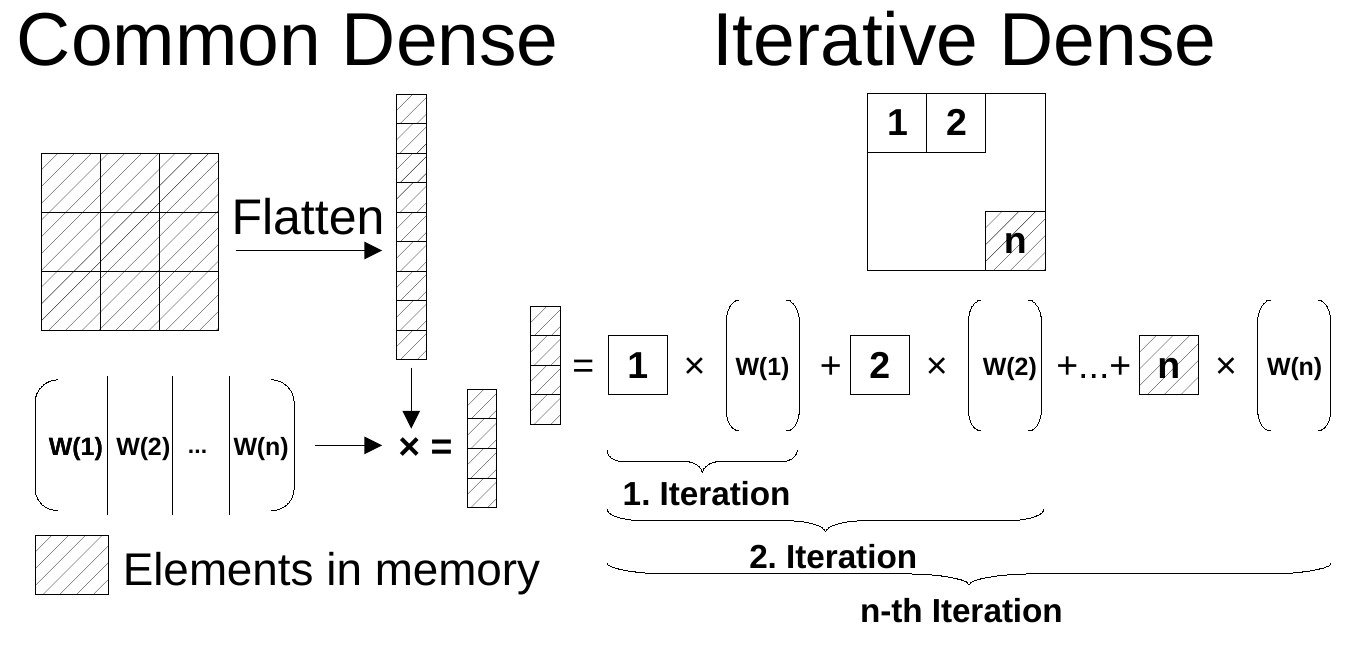}
    \caption{Comparison of common and iterative dense layer. The columns of the weight matrix are denoted as $w(n)$.}
    \label{fig:common_vs_iter_dense}
  \end{minipage}
\end{figure}

\textbf{Iterative Computation of Global Pooling -- }
As illustrated in \cref{fig:common_vs_iter_pool}, standard global pooling requires that all elements of the input tensor stored in \ac{RAM}. In our approach, the global pooling layer receives one or a few input elements at each step and iteratively updates the result. For a $7 \times 7$ global pooling layer, this allows us to compress the \ac{RAM} usage to 2\% of the original size, without introducing any redundant computations or computation overhead.


\textbf{Iterative Computation of Dense Layer -- }
We noted that the matrix multiplication in dense layers can be implemented by splitting the input vector into individual elements, multiplying each element with its corresponding weight column, and iteratively summing the results, as shown in \cref{fig:common_vs_iter_dense}. Unlike the original approach, which requires the complete input tensor, this method processes only one element of the input tensor per iteration. For a 1024→256 dense layer, this approach compresses memory usage to 20\% of the original.


\section{Experiments on Microcontrollers}
\label{sec:experiments}
In this section we report on experiments running msf-CNN on various \acp{MCU}, aiming to validate both the correctness of our optimization strategies and their versatility when applied on diverse hardware. 

More concretely, we measured peak \ac{RAM} usage and compute latency based on the fusion settings in \cref{sec:analy_rslt}, as reported in the following. As shown in \cref{tbl:iot_boards}, we carried out our experiments on the relevant 32-bit microcontroller architectures: Arm Cortex-M, Espressif Xtensa, and RISC-V.
For our model zoo, we chose MBV2-w0.35, MN2-vww5 and MN2-320K as they are good representatives of backbones for applications in \ac{AIoT} \cite{saha2022machine}, as also used in prior works \cite{lin_memory-efficient_2021, zheng2024streamnet}. We compare msf-CNN performance to the closest related work: MCUNetV2~\cite{lin_memory-efficient_2021} and StreamNet-2D~\cite{zheng2024streamnet}, more simply denoted StreamNet in the following.

\begin{table}[ht]
	\centering
	\caption{The different microcontrollers \& boards used in our experiments. The RAM and Flash capacity are presented in kB.}
	\label{tbl:iot_boards}
	\begin{tabular}{lllll}
		\hline
		Board            & MCU                 & Core             & RAM & Flash \\ \hline
		Nucleo-f767zi    & STM32F767ZI         & Cortex-M7 @ 216 MHz     & 512      & 2048       \\
		Stm32f746g-disco & STM32F746NG         & Cortex-M7 @ 216 MHz     & 320      & 1024       \\
		Nucleo-f412zg    & STM32F412ZG         & Cortex-M4 @ 100 MHz     & 256      & 1024       \\
		esp32s3-devkit   & ESP32-S3-WROOM-1N8  & Xtensa @ 240 MHz & 512      & 8192       \\
		esp32c3-devkit   & ESP32c3-1-MINI-M4N4 & RISC-V @ 160 MHz & 384      & 4096       \\
		hifive1b         & SiFive FE310-G002   & RISC-V @ 320 MHz & 16       & 4096       \\ \hline
	\end{tabular}
\end{table}

\subsection{Minimal Peak RAM Usage}
First, we evaluated solutions to P1 while relaxing \cref{equ:mac_limit}, i.e. the fusion settings with minimum peak \ac{RAM} usage, without considering time constraint. Results are shown in \cref{tbl:min_peak_mem}. We observe that, compared to prior art (StreamNet-2D and MCUNetV2), msf-CNN can further reduce the peak \ac{RAM} usage by 65\% to 87\%. We could even deploy the MBV2-w0.35 model onto the SiFive board that provides only 16 kB \ac{RAM} (!). However, achieving this high compression ratio comes at the expense of increased computational latency, which we measured in \cref{tbl:latency-min_peak_mem}. Interestingly, while clock frequency plays a decisive role, MCU architecture can also have a crucial effect, for larger models. For instance, notice latency with Xtensa esp32s3 at 240MHz \emph{versus} RISC-V esp32c3 at 160 MHz, for the MN2-320K model (in \cref{tbl:latency-min_peak_mem}). Nevertheless, we measured that latency increases $2 \times$ to $5 \times$ compared to vanilla (non-fused) CNN. Hence, such minimal RAM settings are only suitable for latency-tolerant applications on the smallest devices.

\begin{table}[htbp]
  \centering
  \begin{minipage}[t]{0.48\linewidth}
    \centering

    \caption{Minimal peak RAM use, measured in kB. (Vanilla: un-fused model)}
	\label{tbl:min_peak_mem}
    \sisetup{table-format = 2.2, round-mode = places, round-precision = 2}
	\begin{tabular}{lSSS}
		\hline
        \textit{(Fusion)} & {\makecell{MBV2\\-w0.35}} & {\makecell{MN2\\-vww5}} & {\makecell{MN2\\-320K}}
         \\\hline
		Vanilla        & 194.44     & 96       & 309.76   \\
		MCUNet\tiny{\textbf{V2}}      & 63         & 45       & 215      \\
		StreamNet  & 66         & 44       & 208      \\
		\textbf{msf-CNN} & 8.56       & 15.368   & 51.164   \\ \hline
	\end{tabular}
  \end{minipage}
  \hfill
  \begin{minipage}[t]{0.48\linewidth}
    \centering
	\caption{Inference execution time, measured in \emph{ms}, with msf-CNN tuned with minimal peak RAM. (OOM: Out-of-Memory)}
	\label{tbl:latency-min_peak_mem}
	\begin{tabular}{lrrr}
		\hline
        \textit{(MCU)} & {\makecell{MBV2\\-w0.35}} & {\makecell{MN2\\-vww5}} & {\makecell{MN2\\-320K}} \\ \hline
        stm32f767 & 1996.8     & 1723.0   & 19329.9  \\
        (vs. vanilla) & $2.5 \times$ & $3.4 \times$ & $4.4 \times$\\ \hline
		stm32f746     & 1379.6     & 1727.5   & 16261.9  \\
		stm32f412     & 5270.1     & 4943.4   & 56979.0  \\
        esp32s3       & 6748.2     & 5974.1   & 76763.6  \\
		esp32c3       & 6792.7     & 6248.9   & 73713.8  \\
		SiFive      & 10000.0    & OOM      & OOM      \\ \hline
	\end{tabular}
  \end{minipage}
\end{table}

\subsection{Impact of \ac{RAM} Budget Limit}

\begin{figure}
    \centering
    \includegraphics[width=\linewidth]{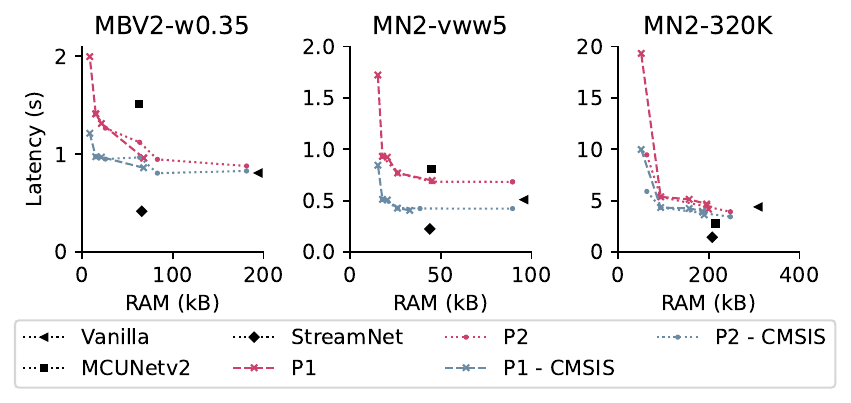}
    \caption{Trade-off between RAM and latency of different optimal fusion settings on Nucleo-f767zi. \added{P1: Minimize RAM s.t. compute cost limit. P2: Minimize compute cost s.t. RAM limit. For more detailed results, please refer to \cref{tbl:impact_mem_and_mac} in \cref{appdix:exp_details}.}}
    \label{fig:ram-latency-trade-off}
\end{figure}

As shown in \cref{fig:ram-latency-trade-off} and \cref{tbl:impact_mem_and_mac}, the measured peak \ac{RAM} usage consistently obeys to the given constraints, thereby validating the correctness of the optimizer and corroborating our analytical results.  Based on these, we observe that higher \ac{RAM} budgets result in shorter compute latency for the optimal fusion configurations identified by msf-CNN. This is because the optimizer tends to favor configurations with either no fusion or shallow fusion depths, which correspond to higher peak \ac{RAM} usage but lower computational costs.

For the MBV2-w0.35 and MN2-vww5 models, our method outperforms MCUNetV2 when the \ac{RAM} limit is set to 32kB and 64kB. Although our method does not surpass StreamNet-2D across the board, msf-CNN does demonstrate its flexibility, enabling users to select the optimal fusion configuration under varying memory budgets.

\subsection{Impact of Computation Cost Limit}
When capping computation cost as a constraint, the relation between compute latency and peak \ac{RAM} usage is consistent (dual) with the previous section, such that higher compute overhead budgets result in longer compute latency and smaller peak \ac{RAM} usage. We also observe that the ratio $F$ measuring the overhead compared to vanilla CNN (no fusion) is bigger than the $F_{max}$ we set for. This discrepancy comes from the fact that the optimizer computes the amount of MAC operations, whereas the full latency includes not only MAC operations but also I/O delays. In mainstream \acp{MCU}, model weights are stored in Flash rather than RAM, which introduces substantial additional latency during read operations, thereby contributing to higher compute latency. Specifically, when recomputation occurs, the weights must be refetched from flash memory, which could disrupt cache hits and lead to increased overall latency.
 Despite this discrepancy, our method still generates fusion configurations for the MBV2-w0.35 and MN2-vww5 models that outperform MCUNetV2. Particularly for memory-sensitive but time-insensitive applications, we can set the constraint $F_{max}$ to infinity, thereby obtaining novel fusion configurations with minimal \ac{RAM} usage.

 msf-CNN underperforms StreamNet because our implementation does not use hardware-dependent libraries or acceleration instructions, while StreamNet is optimized for ARM platforms via the CMSIS \cite{lai2018cmsis} and employs a 2-D cache (ours uses 1-D), which further mitigates recomputation. To enable a fairer comparison, we added an optional CMSIS backend to msf-CNN for ARM devices and conducted additional measurements, as shown in \cref{fig:ram-latency-trade-off}. The results show msf-CNN’s Pareto curves of CMSIS variants are much closer to StreamNet’s. We are also extending msf-CNN with 2-D cache support to provide greater flexibility across hardware platforms.

\section{Discussion}
\label{appdix:limitations}

Our experiments demonstrate msf-CNN's capability to optimize resource usage with diverse CNN models, under user-specified constraints emphasizing either 
compute latency or \ac{RAM} footprint. Furthermore, msf-CNN generates code that is deployable across diverse MCUs' ISAs. Users can thus produce optimal \ac{CNN} fusion configurations tailored to specific industrial hardware requirements. However, some limitations remain, on which our future work will focus next: 

\textbf{Parameter Space -- }
The current optimization scope is limited to fusion block positioning and depth selection, with the number of output elements per iteration fixed at one. This parameter significantly impacts both memory footprint and compute overhead, which warrants further exploration.

\textbf{Caching Paradigm -- }
The search space currently incorporates only the H-cache paradigm. Future implementations should integrate alternative caching strategies to enhance optimization flexibility.

\textbf{Neural Network Architecture -- }
The work currently focuses exclusively on convolutional neural network architectures (CNNs). The analysis of other prevalent structures, particularly attention mechanisms and recurrent neural networks (RNNs), remains an open research direction. Please check \cref{appdix:nn_archi_ext} for the current state of extension.

\section{Related Work}


\textbf{Machine Learning Compilers for MCUs -- }
Compilers such as \ac{TVM}\cite{chen2018tvm}, IREE\cite{The_IREE_Authors_IREE_2019}, FlexTensor~\cite{zheng2020flextensor}, and Buddy~\cite{zhang2023compiler} offer automated transpilation and compilation for models produced by major \ac{ML} frameworks, including TensorFlow and PyTorch. 
Other prior work such as RIOT-ML~\cite{huang2024riot-ml} 
combine a small general-purpose OS with microTVM (extension of TVM orienting to MCU), for comprehensive support for \ac{ML} frameworks and operator implementation on divers \acp{MCU}. 
However, none of the above tools provide CNN fusion optimization mechanisms, in contrast to msf-CNN.

\textbf{Efficient Neural Network Structure -- }
For models to operate on low-power IoT devices, they must be compact and computationally efficient. Studies have demonstrated the use of lightweight CNNs for speech recognition and age classification \cite{maayah_limitaccess_2023}, water leakage detection \cite{atanane_smart_2023}, fall detection for the elderly \cite{fang_fall_2021} and other tasks \cite{hussain_swishnet_2018, zhu-zhou_computationally_2023}. Tiny vision transformers have also been employed for classification tasks in various studies \cite{jinyang_yu_tiny_2023, liang_mcuformer_2023, yao_cnn-transformer_2023, wyatt_environmental_2021}. Besides handcrafting a lightweight structure by reducing layer number or kernel size, people~\cite{iandola_squeezenet_2016, tan_efficientnet_2020, howard_mobilenets_2017, sandler_mobilenetv2_2018, howard_searching_2019} also re-designed the basic blocks to replace common convolutions for lower memory footprint and compute latency.

\textbf{Tiny Neural Architecture Search (NAS) -- } This technique is employed to automatically search for model structures with optimal accuracy under the constraints of memory, flash footprint and compute latency. TinyNAS \cite{lin2020mcunet_tinynas}, $\mu$NAS \cite{liberis2021munas} and the Once-for-All Network \cite{cai2019once} leverage \ac{NAS} to design \acp{CNN} with exceptionally small memory requirements for \acp{MCU}. The resulting networks require only a few hundred kilobytes of RAM for execution. However, contrary to msf-CNN, these methods necessitate retraining or fine-tuning of pre-existing networks.

\textbf{Memory Optimization for CNN layers -- }
Memory optimization strategies can be broadly categorized into scheduling-based and fusion-based methods. Scheduling-based methods, such as those implemented in frameworks like \added{Ansor~\cite{zheng2020ansor}}, vMCU~\cite{zheng_vmcu_2024}, MoDEL \cite{steiner2023model} and  TinyEngine~\cite{lin_memory-efficient_2021}, focus on the efficient reuse of memory pools to minimize peak memory usage by leveraging the different lifetimes of inter- and intra-layer tensors. 
Although both methods achieve a peak memory reduction exceeding 50\%, they still generate a complete output tensor for each layer. This requirement remains problematic for low-power \acp{MCU} with limited \ac{RAM}. 
Prior work on fusion was covered in Section~\ref{sec:background}. Contrary to msf-CNN, these methods do not fully exploit the potential of multiple fusion blocks.

\section{Conclusion}

Convolutional neural networks (CNNs) must not only execute in the cloud or on edge computing gateways,
but also on the smaller, more energy-efficient microcontroller-based devices which take part in our cyber-physical systems.
Microcontrollers pose a great challenge  for CNNs regarding the
joint optimization of RAM memory consumption and inference latency.
In this context, we presented msf-CNN, a technique and heuristics
able to identify pools of practical patch-based fusion optimizations for CNN inference, which jointly satisfy
memory and latency constraints. Compared to previous work on CNN fusion for microcontrollers, msf-CNN
identifies a wider set of applicable solutions, on much more diverse hardware.
Our experimental evaluation using the open source implementation we provide for common microcontrollers (ARM Cortex-M, RISC-V, and ESP32)
show that msf-CNN can achieve inference with less than 50\% the peak RAM usage compared to state-of-the-art.
As such, msf-CNN provides a new level of flexibility for embedded system designers, which can now better tune the trade-off between peak RAM and model
inference latency on various hardware.

\section*{Acknowledgment}
The authors thank Adrien Tousnakhoff for useful discussions. Work contributing to these results was partially funded
by the ANR France 2030 Programme (ANR-22-PTCC-0001 and ANR-22-PEFT-0007).

\bibliography{citation}

\begin{thebibliography}{50}
\providecommand{\natexlab}[1]{#1}
\providecommand{\url}[1]{\texttt{#1}}
\expandafter\ifx\csname urlstyle\endcsname\relax
  \providecommand{\doi}[1]{doi: #1}\else
  \providecommand{\doi}{doi: \begingroup \urlstyle{rm}\Url}\fi

\bibitem[Abadi et~al.(2016)Abadi, Barham, Chen, Chen, Davis, Dean, Devin, Ghemawat, Irving, Isard, et~al.]{abadi2016tensorflow}
Abadi, M., Barham, P., Chen, J., Chen, Z., Davis, A., Dean, J., Devin, M., Ghemawat, S., Irving, G., Isard, M., et~al.
\newblock $\{$TensorFlow$\}$: a system for $\{$Large-Scale$\}$ machine learning.
\newblock In \emph{12th USENIX symposium on operating systems design and implementation (OSDI 16)}, pp.\  265--283, 2016.

\bibitem[Alwani et~al.(2016)Alwani, Chen, Ferdman, and Milder]{alwani_fused-layer_2016}
Alwani, M., Chen, H., Ferdman, M., and Milder, P.
\newblock Fused-layer {CNN} accelerators.
\newblock In \emph{2016 49th {Annual} {IEEE}/{ACM} {International} {Symposium} on {Microarchitecture} ({MICRO})}, pp.\  1--12, Taipei, Taiwan, October 2016. IEEE.
\newblock ISBN 978-1-5090-3508-3.
\newblock \doi{10.1109/MICRO.2016.7783725}.
\newblock URL \url{http://ieeexplore.ieee.org/document/7783725/}.

\bibitem[Atanane et~al.(2023)Atanane, Mourhir, Benamar, and Zennaro]{atanane_smart_2023}
Atanane, O., Mourhir, A., Benamar, N., and Zennaro, M.
\newblock Smart {Buildings}: {Water} {Leakage} {Detection} {Using} {TinyML}.
\newblock \emph{Sensors}, 23\penalty0 (22):\penalty0 9210, November 2023.
\newblock ISSN 1424-8220.
\newblock \doi{10.3390/s23229210}.
\newblock URL \url{https://www.mdpi.com/1424-8220/23/22/9210}.

\bibitem[Baccelli et~al.(2018)Baccelli, G{\"u}ndo{\u{g}}an, Hahm, Kietzmann, Lenders, Petersen, Schleiser, Schmidt, and W{\"a}hlisch]{baccelli2018riot}
Baccelli, E., G{\"u}ndo{\u{g}}an, C., Hahm, O., Kietzmann, P., Lenders, M.~S., Petersen, H., Schleiser, K., Schmidt, T.~C., and W{\"a}hlisch, M.
\newblock {RIOT: An Open Source Operating System for Low-end Embedded Devices in the IoT}.
\newblock \emph{IEEE Internet of Things Journal}, 5\penalty0 (6):\penalty0 4428--4440, 2018.

\bibitem[Bormann et~al.(2014)]{rfc7228}
Bormann, C. et~al.
\newblock {Terminology for Constrained-Node Networks}.
\newblock RFC 7228, May 2014.

\bibitem[Cai et~al.(2019)Cai, Gan, Wang, Zhang, and Han]{cai2019once}
Cai, H., Gan, C., Wang, T., Zhang, Z., and Han, S.
\newblock Once-for-all: Train one network and specialize it for efficient deployment.
\newblock \emph{arXiv preprint arXiv:1908.09791}, 2019.

\bibitem[Chen et~al.(2018)Chen, Moreau, Jiang, Zheng, Yan, Cowan, Shen, Wang, Hu, Ceze, et~al.]{chen2018tvm}
Chen, T., Moreau, T., Jiang, Z., Zheng, L., Yan, E., Cowan, M., Shen, H., Wang, L., Hu, Y., Ceze, L., et~al.
\newblock Tvm: An automated end-to-end optimizing compiler for deep learning.
\newblock \emph{arXiv preprint arXiv:1802.04799}, 2018.

\bibitem[Dijkstra(2022)]{dijkstra2022note}
Dijkstra, E.~W.
\newblock A note on two problems in connexion with graphs.
\newblock In \emph{Edsger Wybe Dijkstra: his life, work, and legacy}, pp.\  287--290. 2022.

\bibitem[Fang et~al.(2021)Fang, Xu, Li, and Pan]{fang_fall_2021}
Fang, K., Xu, Z., Li, Y., and Pan, J.
\newblock A {Fall} {Detection} using {Sound} {Technology} {Based} on {TinyML}.
\newblock In \emph{2021 11th {International} {Conference} on {Information} {Technology} in {Medicine} and {Education} ({ITME})}, pp.\  222--225, Wuyishan, Fujian, China, November 2021. IEEE.
\newblock ISBN 978-1-66540-679-6.
\newblock \doi{10.1109/ITME53901.2021.00053}.
\newblock URL \url{https://ieeexplore.ieee.org/document/9750658/}.

\bibitem[Fredman \& Tarjan(1987)Fredman and Tarjan]{fredman1987fibonacci}
Fredman, M.~L. and Tarjan, R.~E.
\newblock Fibonacci heaps and their uses in improved network optimization algorithms.
\newblock \emph{Journal of the ACM (JACM)}, 34\penalty0 (3):\penalty0 596--615, 1987.

\bibitem[Ghosh et~al.(2018)Ghosh, Chakraborty, and Law]{ghosh2018artificial}
Ghosh, A., Chakraborty, D., and Law, A.
\newblock Artificial intelligence in internet of things.
\newblock \emph{CAAI Transactions on Intelligence Technology}, 3\penalty0 (4):\penalty0 208--218, 2018.

\bibitem[Guo et~al.(2022)Guo, Lu, Hou, Liu, Cheng, and Hu]{guo2022segnext}
Guo, M.-H., Lu, C.-Z., Hou, Q., Liu, Z., Cheng, M.-M., and Hu, S.-M.
\newblock Segnext: Rethinking convolutional attention design for semantic segmentation.
\newblock \emph{Advances in neural information processing systems}, 35:\penalty0 1140--1156, 2022.

\bibitem[Guo et~al.(2023)Guo, Lu, Liu, Cheng, and Hu]{guo2023visual}
Guo, M.-H., Lu, C.-Z., Liu, Z.-N., Cheng, M.-M., and Hu, S.-M.
\newblock Visual attention network.
\newblock \emph{Computational visual media}, 9\penalty0 (4):\penalty0 733--752, 2023.

\bibitem[He et~al.(2016)He, Zhang, Ren, and Sun]{he2016deepresnet}
He, K., Zhang, X., Ren, S., and Sun, J.
\newblock Deep residual learning for image recognition.
\newblock In \emph{Proceedings of the IEEE conference on computer vision and pattern recognition}, pp.\  770--778, 2016.

\bibitem[Howard et~al.({\natexlab{a}})Howard, Sandler, Chu, Chen, Chen, Tan, Wang, Zhu, Pang, Vasudevan, Le, and Adam]{howard_searching_2019}
Howard, A., Sandler, M., Chu, G., Chen, L.-C., Chen, B., Tan, M., Wang, W., Zhu, Y., Pang, R., Vasudevan, V., Le, Q.~V., and Adam, H.
\newblock Searching for {MobileNetV}3.
\newblock pp.\  1314--1324, {\natexlab{a}}.
\newblock URL \url{https://openaccess.thecvf.com/content_ICCV_2019/html/Howard_Searching_for_MobileNetV3_ICCV_2019_paper.html}.

\bibitem[Howard et~al.({\natexlab{b}})Howard, Zhu, Chen, Kalenichenko, Wang, Weyand, Andreetto, and Adam]{howard_mobilenets_2017}
Howard, A.~G., Zhu, M., Chen, B., Kalenichenko, D., Wang, W., Weyand, T., Andreetto, M., and Adam, H.
\newblock {MobileNets}: Efficient convolutional neural networks for mobile vision applications, {\natexlab{b}}.
\newblock URL \url{http://arxiv.org/abs/1704.04861}.

\bibitem[Huang et~al.(2024)Huang, Zandberg, Schleiser, and Baccelli]{huang2024riot-ml}
Huang, Z., Zandberg, K., Schleiser, K., and Baccelli, E.
\newblock {RIOT-ML: toolkit for over-the-air secure updates and performance evaluation of TinyML models}.
\newblock \emph{Annals of Telecommunications}, pp.\  1--15, 2024.

\bibitem[Hussain \& Haque(2018)Hussain and Haque]{hussain_swishnet_2018}
Hussain, M.~S. and Haque, M.~A.
\newblock {SwishNet}: {A} {Fast} {Convolutional} {Neural} {Network} for {Speech}, {Music} and {Noise} {Classification} and {Segmentation}.
\newblock 2018.
\newblock \doi{10.48550/ARXIV.1812.00149}.
\newblock URL \url{https://arxiv.org/abs/1812.00149}.

\bibitem[Iandola et~al.()Iandola, Han, Moskewicz, Ashraf, Dally, and Keutzer]{iandola_squeezenet_2016}
Iandola, F.~N., Han, S., Moskewicz, M.~W., Ashraf, K., Dally, W.~J., and Keutzer, K.
\newblock {SqueezeNet}: {AlexNet}-level accuracy with 50x fewer parameters and{\textless} 0.5 {MB} model size.

\bibitem[{Jinyang Yu} et~al.(2023){Jinyang Yu}, {Zikai Song}, {Jiahao Ji}, {Lixian Zhu}, {Kele Xu}, Qian, Dou, and Hu]{jinyang_yu_tiny_2023}
{Jinyang Yu}, {Zikai Song}, {Jiahao Ji}, {Lixian Zhu}, {Kele Xu}, Qian, K., Dou, Y., and Hu, B.
\newblock Tiny {Audio} {Spectrogram} {Transformer}: {Mobilevit} for {Low}-{Complexity} {Acoustic} {Scene} {Classification} with {Decoupled} {Knowledge} {Distillation}.
\newblock 2023.
\newblock \doi{10.13140/RG.2.2.24001.12646}.
\newblock URL \url{https://rgdoi.net/10.13140/RG.2.2.24001.12646}.

\bibitem[Koonce(2021)]{koonce2021resnet-34}
Koonce, B.
\newblock Resnet 34.
\newblock \emph{Convolutional neural networks with swift for tensorflow: image recognition and dataset categorization}, pp.\  51--61, 2021.

\bibitem[Lai et~al.(2018)Lai, Suda, and Chandra]{lai2018cmsis}
Lai, L., Suda, N., and Chandra, V.
\newblock Cmsis-nn: Efficient neural network kernels for arm cortex-m cpus.
\newblock \emph{arXiv preprint arXiv:1801.06601}, 2018.

\bibitem[Liang et~al.(2023)Liang, Wang, Xu, Tang, Zhou, and Lu]{liang_mcuformer_2023}
Liang, Y., Wang, Z., Xu, X., Tang, Y., Zhou, J., and Lu, J.
\newblock {MCUFormer}: {Deploying} {Vision} {Transformers} on {Microcontrollers} with {Limited} {Memory}.
\newblock 2023.
\newblock \doi{10.48550/ARXIV.2310.16898}.
\newblock URL \url{https://arxiv.org/abs/2310.16898}.

\bibitem[Liberis et~al.(2021)Liberis, Dudziak, and Lane]{liberis2021munas}
Liberis, E., Dudziak, {\L}., and Lane, N.~D.
\newblock $\mu$nas: Constrained neural architecture search for microcontrollers.
\newblock In \emph{Proceedings of the 1st Workshop on Machine Learning and Systems}, pp.\  70--79, 2021.

\bibitem[Lin et~al.(2020)Lin, Chen, Lin, Gan, Han, et~al.]{lin2020mcunet_tinynas}
Lin, J., Chen, W.-M., Lin, Y., Gan, C., Han, S., et~al.
\newblock Mcunet: Tiny deep learning on iot devices.
\newblock \emph{Advances in neural information processing systems}, 33:\penalty0 11711--11722, 2020.

\bibitem[Lin et~al.(2021)Lin, Chen, Cai, Gan, and Han]{lin_memory-efficient_2021}
Lin, J., Chen, W.-M., Cai, H., Gan, C., and Han, S.
\newblock Memory-efficient {Patch}-based {Inference} for {Tiny} {Deep} {Learning}.
\newblock In \emph{Advances in {Neural} {Information} {Processing} {Systems}}, volume~34, pp.\  2346--2358. Curran Associates, Inc., 2021.
\newblock URL \url{https://proceedings.neurips.cc/paper/2021/hash/1371bccec2447b5aa6d96d2a540fb401-Abstract.html}.

\bibitem[Maayah et~al.(2023)Maayah, Abunada, Al-Janahi, Ahmed, and Qadir]{maayah_limitaccess_2023}
Maayah, M., Abunada, A., Al-Janahi, K., Ahmed, M.~E., and Qadir, J.
\newblock {LimitAccess}: on-device {TinyML} based robust speech recognition and age classification.
\newblock \emph{Discover Artificial Intelligence}, 3\penalty0 (1):\penalty0 8, February 2023.
\newblock ISSN 2731-0809.
\newblock \doi{10.1007/s44163-023-00051-x}.
\newblock URL \url{https://link.springer.com/10.1007/s44163-023-00051-x}.

\bibitem[Mei et~al.(2023)Mei, Goetschalckx, Symons, and Verhelst]{mei2023defines}
Mei, L., Goetschalckx, K., Symons, A., and Verhelst, M.
\newblock Defines: Enabling fast exploration of the depth-first scheduling space for dnn accelerators through analytical modeling.
\newblock In \emph{2023 IEEE International Symposium on High-Performance Computer Architecture (HPCA)}, pp.\  570--583. IEEE, 2023.

\bibitem[Niu et~al.(2021)Niu, Guan, Wang, Agrawal, and Ren]{niu2021dnnfusion}
Niu, W., Guan, J., Wang, Y., Agrawal, G., and Ren, B.
\newblock Dnnfusion: accelerating deep neural networks execution with advanced operator fusion.
\newblock In \emph{Proceedings of the 42nd ACM SIGPLAN International Conference on Programming Language Design and Implementation}, pp.\  883--898, 2021.

\bibitem[Paszke et~al.(2019)Paszke, Gross, Massa, Lerer, Bradbury, Chanan, Killeen, Lin, Gimelshein, Antiga, et~al.]{paszke2019pytorch}
Paszke, A., Gross, S., Massa, F., Lerer, A., Bradbury, J., Chanan, G., Killeen, T., Lin, Z., Gimelshein, N., Antiga, L., et~al.
\newblock Pytorch: An imperative style, high-performance deep learning library.
\newblock \emph{Advances in neural information processing systems}, 32, 2019.

\bibitem[Pinckaers et~al.(2022)Pinckaers, van Ginneken, and Litjens]{pinckaers_streaming_2022}
Pinckaers, H., van Ginneken, B., and Litjens, G.
\newblock Streaming {Convolutional} {Neural} {Networks} for {End}-to-{End} {Learning} {With} {Multi}-{Megapixel} {Images}.
\newblock \emph{IEEE Transactions on Pattern Analysis and Machine Intelligence}, 44\penalty0 (3):\penalty0 1581--1590, March 2022.
\newblock ISSN 1939-3539.
\newblock \doi{10.1109/TPAMI.2020.3019563}.
\newblock URL \url{https://ieeexplore.ieee.org/document/9178453}.
\newblock Conference Name: IEEE Transactions on Pattern Analysis and Machine Intelligence.

\bibitem[Robert(2002)]{robert2002algorithms}
Robert, S.
\newblock Algorithms in c, part 5: Graph algorithms, 2002.

\bibitem[Saha et~al.(2022)Saha, Sandha, and Srivastava]{saha2022machine}
Saha, S.~S., Sandha, S.~S., and Srivastava, M.
\newblock Machine learning for microcontroller-class hardware: A review.
\newblock \emph{IEEE Sensors Journal}, 22\penalty0 (22):\penalty0 21362--21390, 2022.

\bibitem[Sandler et~al.()Sandler, Howard, Zhu, Zhmoginov, and Chen]{sandler_mobilenetv2_2018}
Sandler, M., Howard, A., Zhu, M., Zhmoginov, A., and Chen, L.-C.
\newblock {MobileNetV}2: Inverted residuals and linear bottlenecks.
\newblock pp.\  4510--4520.
\newblock URL \url{https://openaccess.thecvf.com/content_cvpr_2018/html/Sandler_MobileNetV2_Inverted_Residuals_CVPR_2018_paper.html}.

\bibitem[Sedgewick(2001)]{sedgewik2001graph_algorithms}
Sedgewick, R.
\newblock \emph{Algorithms in c, part 5: graph algorithms, third edition}.
\newblock Addison-Wesley Professional, third edition, 2001.
\newblock ISBN 9780768685329.

\bibitem[Shen et~al.(2021)Shen, Zhang, Zhao, Yi, and Li]{shen2021efficient}
Shen, Z., Zhang, M., Zhao, H., Yi, S., and Li, H.
\newblock Efficient attention: Attention with linear complexities.
\newblock In \emph{Proceedings of the IEEE/CVF winter conference on applications of computer vision}, pp.\  3531--3539, 2021.

\bibitem[Steiner et~al.(2023)Steiner, Elhoushi, Kahn, and Hegarty]{steiner2023model}
Steiner, B., Elhoushi, M., Kahn, J., and Hegarty, J.
\newblock Model: memory optimizations for deep learning.
\newblock In \emph{International Conference on Machine Learning}, pp.\  32618--32632. PMLR, 2023.

\bibitem[Tan \& Le()Tan and Le]{tan_efficientnet_2020}
Tan, M. and Le, Q.~V.
\newblock {EfficientNet}: Rethinking model scaling for convolutional neural networks.
\newblock URL \url{http://arxiv.org/abs/1905.11946}.

\bibitem[{The IREE Authors}(2019)]{The_IREE_Authors_IREE_2019}
{The IREE Authors}.
\newblock {IREE}, September 2019.
\newblock URL \url{https://github.com/iree-org/iree}.

\bibitem[Wang et~al.(2010)Wang, Lin, and Yi]{wang2010kernel}
Wang, G., Lin, Y., and Yi, W.
\newblock Kernel fusion: An effective method for better power efficiency on multithreaded gpu.
\newblock In \emph{2010 IEEE/ACM Int'l Conference on Green Computing and Communications \& Int'l Conference on Cyber, Physical and Social Computing}, pp.\  344--350. IEEE, 2010.

\bibitem[Wyatt et~al.(2021)Wyatt, Elliott, Aravamudan, Otero, Otero, Anagnostopoulos, Smith, Peter, Jones, Leung, and Lam]{wyatt_environmental_2021}
Wyatt, S., Elliott, D., Aravamudan, A., Otero, C.~E., Otero, L.~D., Anagnostopoulos, G.~C., Smith, A.~O., Peter, A.~M., Jones, W., Leung, S., and Lam, E.
\newblock Environmental {Sound} {Classification} with {Tiny} {Transformers} in {Noisy} {Edge} {Environments}.
\newblock In \emph{2021 {IEEE} 7th {World} {Forum} on {Internet} of {Things} ({WF}-{IoT})}, pp.\  309--314, June 2021.
\newblock \doi{10.1109/WF-IoT51360.2021.9596007}.
\newblock URL \url{https://ieeexplore.ieee.org/abstract/document/9596007}.

\bibitem[Xingjian(2015)]{xingjian2015convolutionallstm}
Xingjian, S.
\newblock Convolutional lstm network: A machine learning approach for precipitation nowcasting.
\newblock \emph{Advances in neural information processing systems}, 28:\penalty0 1, 2015.

\bibitem[Yao \& Liu(2023)Yao and Liu]{yao_cnn-transformer_2023}
Yao, Z. and Liu, X.
\newblock A {CNN}-{Transformer} {Deep} {Learning} {Model} for {Real}-time {Sleep} {Stage} {Classification} in an {Energy}-{Constrained} {Wireless} {Device} $^{\textrm{*}}$.
\newblock In \emph{2023 11th {International} {IEEE}/{EMBS} {Conference} on {Neural} {Engineering} ({NER})}, pp.\  1--4, Baltimore, MD, USA, April 2023. IEEE.
\newblock ISBN 978-1-66546-292-1.
\newblock \doi{10.1109/NER52421.2023.10123825}.
\newblock URL \url{https://ieeexplore.ieee.org/document/10123825/}.

\bibitem[Zhang et~al.(2023)Zhang, Xing, Wu, and Zhao]{zhang2023compiler}
Zhang, H., Xing, M., Wu, Y., and Zhao, C.
\newblock Compiler technologies in deep learning co-design: A survey.
\newblock \emph{Intelligent Computing}, 2:\penalty0 0040, 2023.

\bibitem[Zhao et~al.(2022)Zhao, Gao, Xia, Zhang, Chen, Chen, Zhang, Geng, Cheng, and Jin]{zhao2022apollo}
Zhao, J., Gao, X., Xia, R., Zhang, Z., Chen, D., Chen, L., Zhang, R., Geng, Z., Cheng, B., and Jin, X.
\newblock Apollo: Automatic partition-based operator fusion through layer by layer optimization.
\newblock \emph{Proceedings of Machine Learning and Systems}, 4:\penalty0 1--19, 2022.

\bibitem[Zheng et~al.(2024{\natexlab{a}})Zheng, Liu, Hsu, and Yeh]{zheng2024streamnet}
Zheng, H.-S., Liu, Y.-Y., Hsu, C.-F., and Yeh, T.~T.
\newblock Streamnet: memory-efficient streaming tiny deep learning inference on the microcontroller.
\newblock \emph{Advances in Neural Information Processing Systems}, 36, 2024{\natexlab{a}}.

\bibitem[Zheng et~al.(2020{\natexlab{a}})Zheng, Jia, Sun, Wu, Yu, Haj-Ali, Wang, Yang, Zhuo, Sen, et~al.]{zheng2020ansor}
Zheng, L., Jia, C., Sun, M., Wu, Z., Yu, C.~H., Haj-Ali, A., Wang, Y., Yang, J., Zhuo, D., Sen, K., et~al.
\newblock Ansor: Generating $\{$High-Performance$\}$ tensor programs for deep learning.
\newblock In \emph{14th USENIX symposium on operating systems design and implementation (OSDI 20)}, pp.\  863--879, 2020{\natexlab{a}}.

\bibitem[Zheng et~al.(2020{\natexlab{b}})Zheng, Liang, Wang, Chen, and Sheng]{zheng2020flextensor}
Zheng, S., Liang, Y., Wang, S., Chen, R., and Sheng, K.
\newblock Flextensor: An automatic schedule exploration and optimization framework for tensor computation on heterogeneous system.
\newblock In \emph{Proceedings of the Twenty-Fifth International Conference on Architectural Support for Programming Languages and Operating Systems}, pp.\  859--873, 2020{\natexlab{b}}.

\bibitem[Zheng et~al.(2024{\natexlab{b}})Zheng, Chen, Li, Ye, Ceze, and Liang]{zheng_vmcu_2024}
Zheng, S., Chen, R., Li, M., Ye, Z., Ceze, L., and Liang, Y.
\newblock {vMCU}: {Coordinated} {Memory} {Management} and {Kernel} {Optimization} for {DNN} {Inference} on {MCUs}.
\newblock \emph{Proceedings of Machine Learning and Systems}, 6:\penalty0 452--464, May 2024{\natexlab{b}}.

\bibitem[Zhu-Zhou et~al.(2023)Zhu-Zhou, Tejera-Berengué, Gil-Pita, Utrilla-Manso, and Rosa-Zurera]{zhu-zhou_computationally_2023}
Zhu-Zhou, F., Tejera-Berengué, D., Gil-Pita, R., Utrilla-Manso, M., and Rosa-Zurera, M.
\newblock Computationally constrained audio-based violence detection through transfer learning and data augmentation techniques.
\newblock \emph{Applied Acoustics}, 213:\penalty0 109638, October 2023.
\newblock ISSN 0003-682X.
\newblock \doi{10.1016/j.apacoust.2023.109638}.
\newblock URL \url{https://www.sciencedirect.com/science/article/pii/S0003682X2300436X}.

\end{thebibliography}
\bibliographystyle{icml2025}

\newpage
\appendix
\section{Relation between msf-CNN and kernel fusion}
\label{appdix:rel_msf_kernel_fusion}

We note that readers might potentially confuse patch-based multi-stage fusion (msf-CNN, our approach) on the one hand, and on the other hand traditional kernel fusion techniques. These two approaches are orthogonal and can be applied concurrently for maximum benefit.

While kernel fusion optimizes computation overhead, msf-CNN instead targets memory efficiency, the latter being the critical first hurdle on small edge devices. As such, we disambiguate this further:

\textbf{Kernel fusion} \cite{wang2010kernel, niu2021dnnfusion, zhao2022apollo} focuses primarily on reducing redundant data movements between GPU and RAM by combining multiple primitive operators (e.g. Batch Normalization, ReLU, Softmax, etc.) with a primary, memory-bound operator (e.g. conv, pooling) into a single kernel. While kernel fusion improves compute latency and data throughput, it does not address the fundamental memory usage issue that arises when processing multiple primary operators sequentially.

\textbf{Patch-based multi-stage fusion (msf-CNN, our approach)} extends the idea of patch-based fusion \cite{alwani_fused-layer_2016, mei2023defines}. More specifically, msf-CNN:

\begin{itemize}
    \item Fuses multiple layers (i.e. primary operators like convolution and pooling) into a single computational stage.
Implements patch-based partial computation, which drastically reduces peak memory usage by processing input data in smaller patches while maintaining accuracy.
    \item Introduces a compute-memory trade-off mechanism that allows users to prioritize either memory consumption or computational efficiency based on their deployment constraints.
\end{itemize}

This makes msf-CNN fundamentally different from traditional kernel fusion techniques.

To the best of our knowledge, the closest related works to msf-CNN are StreamNet and MCUNetv2, which have been the state-of-the-art for patch-based fusion on microcontrollers so far. Compared to this state of the art, based on our measurements, msf-CNN achieves up to 50\% reduction of peak RAM usage in model inference, for the same inference accuracy, which thus enables more models to fit smaller devices.

\section{Analysis of the H-cache buffer size}
\label{appdix:cache_buffer_size}

For a fusion block containing $n$ layers, the cache buffer size of the $i$-th layer under H-cache scheme is given by
\begin{align} \label{equ:buf_size_ith_layer}
	\text{Buf}_i = t_i \times k_i \times c^{in}_i
\end{align}
where $t_i$, $k_i$ and $c^{in}_i$ are the tile size, kernel size and input channels number, respectively. Obviously, the first layer of the fusion block does not need any input cache, thus $\text{Buf}_1 = 0$. The total cache size of the fusion block is $\text{Buf} = \sum_i \text{Buf}_i$.

\section{Analysis of the amount of MAC operations}
\label{appdix:mac_compute}

Analyzing the number of \ac{MAC} operations in the fusion block is quite complex. The input tensor for each layer is sliced into overlapped tiles, and the kernel performs convolution on the data within each tile. Here, the number of overlapped tiles $N^{tile}$ of each layer is
\begin{align}
	N^{tile} = \lfloor \frac{h^{in} + 2p - t}{s^{tile}} + 1\rfloor \lfloor \frac{w^{in} + 2p - k}{s^{layer}} + 1\rfloor,
\end{align}
where $h^{in}, w^{in}$ are the height and width of input tensor, $s^{tile}, s^{layer}$ are the stride of tile and layer, $p$ represents the input padding. 
Recall that $t$, $k$  are the tile size and kernel size respectively.

And the output size of each tile is determined as:
\begin{align}
	O^{tile} = \lfloor \frac{t - k}{s^{layer}} + 1\rfloor c^{out}.
\end{align}
whereby $c^{out}$ is the number of output channels. We can therefore derive the number $C^{layer}$ of \ac{MAC} operations of a fused convolutional layer as:
\begin{align}\label{equ:mac_fused_layer}
	C^{layer} = N^{tile} \times O^{tile} \times k^2 \times c^{out}.
\end{align}
Finally, we can derive $C^{fb}$ the total \ac{MAC} operations of the entire fusion block as:
\begin{align}\label{equ:mac_fusion_block}
	C^{fb} = \sum C^{layer}.
\end{align}

\section{Complexity analysis of the search algorithm}
\label{sec:complexity_analysis}

We provide a quick preliminary analysis of the \textit{worst-case scenario}. We also highlight that these shortest path computations do not take place on the microcontroller at runtime, but offline on a PC (which expands the realm of what can be assessed as bearable computation).

First, we consider the \textit{lower-bound} of the search algorithm. As shown in Section 6, both Problems P1 and P2 without constraints can be transformed into a multiple single-source-single-target shortest path problem, which can by solved by Dijkstra's algorithm with Fibonacci heap \cite{fredman1987fibonacci} with complexity:

$$
O(E + V log(V)),
$$

where $E$ and $V$ denote the edges (possible fusion blocks) and vertices (layers) of the DAG. In the worst case $E=\sum_{n=1}^V (n-1)$, leading the \textit{overall lower-bound} to $O(V^2)$.

Concerning Problem P1 with constraints: if we don't prune the search space (iteratively), we need to brute force all possible fusion settings of the DAG to form a subspace that fulfills the latency constraints. This leads to enumerating all simple paths from input layer to the output layer. In the worst-case, starting from the input layer, we obtain $2^{V-2}$ fusion combinations, which becomes unbearable for a deep neural network. We can prove this complexity by simple induction:

In worst-case scenario, we assume a complete DAG, where each node is connected to all its predecessors.

\textbf{Base Case:}
For $V = 2$, there is only one complete compute path:

$$
2^{V - 2} = 2^0 = 1.
$$

\textbf{Inductive Step:}
Assume a complete DAG with $k$ nodes has $2^{k - 2}$ complete compute paths.
To construct a complete DAG with $k + 1$ nodes:

\begin{enumerate}
    \item Add a new node with the same incoming edges as the last node. This contributes $2^{k - 2}$ paths.
    \item Connect the new node to the duplicated node, adding another $2^{k - 2}$ paths.
\end{enumerate}

Total paths:

$$
2^{k - 2} + 2^{k - 2} = 2^{k - 1}.
$$

Therefore, by induction, a complete DAG with $V$ nodes has $2^{V - 2}$ complete compute paths. $\blacksquare$

Hence, we apply a pruning strategy (see Equ. 9-11) to reduce the complexity from $O(2^{V-2})$ to $O(V^3)$. The idea: we erase the edges with maximal RAM usage per iteration. In the worst-case, only one edge is erased in each iteration, with a complete DAG with $\frac{V(V-1)}{2}$ edges. Thus, the worst-case complexity of our search algorithm for constrained Problem P1 is $O(V^3)$.

\section{Accuracy Evaluation}
\label{appdix:acc_eval}
msf-CNN is a computation scheduling and memory optimization technique that does not alter the model’s architecture, parameters, or the mathematical operations performed. Like MCUNet and StreamNet, msf-CNN only changes when and how intermediate results are computed and stored to minimize peak memory. Therefore, the final output, and consequently the model’s accuracy, remains identical to the original, unfused model. Hence, standard ML performance benchmarks focusing on accuracy are irrelevant here.

Nevertheless, to make sure, we conducted extra experiments on imagenet and vww dataset, as MCUNet did. We reused the model weights from the pre-trained MCUNet models, and selected $P_{max}=64kB$ as fusion constraints. The results show in \cref{tbl:acc_eval} that no Top-1 accuracy drop was found between vanilla models and the msf-CNN variants.

\begin{table}[]
\centering
\caption{Model Top-1 Accuracy}
\label{tbl:acc_eval}
\begin{tabular}{lrrr}
\hline
Method  & MBV2-w0.35 & MN2-vww5 & MN2-320K \\\hline
Vanilla & 48.94~\%    & 88.89~\%  & 61.76~\%  \\
msf-CNN & 48.94~\%    & 88.89~\%  & 61.76~\%  \\\hline
\end{tabular}
\end{table}

\section{Experiment Details}
\label{appdix:exp_details}

Here, we provide additional details of the experiments, including results not presented in the main text.

\subsection{Supplemental Results}

\cref{tbl:impact_mem_and_mac} provides the underlying data corresponding to \cref{fig:ram-latency-trade-off}, including the specific constraints applied to P1 and P2, respectively.

\begin{table}[htbp]
	\centering
	\caption{Optimal fusion settings on Nucleo-f767zi. RAM (kB), Latency (ms). SAA: Same as above. \colorbox{lightgray}{Gray: msf-CNN beats MCUNet\tiny{V2}.} }
	\label{tbl:impact_mem_and_mac}
\sisetup{table-format = 2.2, round-mode = places, round-precision = 2, table-space-text-post = {$^{***}$}, table-number-alignment = right, table-alignment-mode = none}
\begin{tabular}{@{}
                ll
                S
                S
                S
                S
                S
                S
                @{}}
	\hline
	&         & \multicolumn{2}{l}{MBV2-w0.35}    & \multicolumn{2}{l}{MN2-vww5}     & \multicolumn{2}{l}{MN2-320K}                       \\ \hline
	&         & {RAM}             & {Latency}         & {RAM}             & {Latency}        & {RAM}                      & {Latency}                 \\ \hline
	\multicolumn{2}{l}{Vanilla}       & 194.44          & 807.6           & 96              & 509.7          & 309.76                   & 4394.3                  \\
	\multicolumn{2}{l}{MCUNet\tiny{\textbf{V2}}}     & 63              & 1513.0          & 45              & 810.0          & 215                      & 2777.0                  \\
	\multicolumn{2}{l}{StreamNet} & 66              & 417.0           & 44              & 225.0          & 208                      & 1444.0                  \\ \hline
	\multicolumn{8}{c}{P1: Min. RAM s.t. Compute Cost Limit}                                                                                                   \\\hline
	\multirow{6}{*}{$F_{max}$}  & 1.1     & 67.996          & 961.9           & \gray{45.283} & \gray{696.0} & 199.6                    & 4171.0                  \\
	& 1.2     & \multicolumn{2}{c}{(SAA)}         & \gray{26.24}  & \gray{769.2} & 196.072                  & 4525.1                  \\
	& 1.3     & \gray{21.389} & \gray{1313.8} & 20.568          & 922.7          & 195.333                  & 4680.7                  \\
	& 1.4     & \gray{15.199} & \gray{1412.3} & 17.904          & 931.3          & 156.864                  & 5128.9                  \\
	& 1.5     & \multicolumn{2}{c}{(SAA)}         & \multicolumn{2}{c}{(SAA)}        & 94.224                   & 5370.3                  \\
	& Inf     & 8.56            & 1996.8          & 15.368          & 1723.0         & 51.164                   & 19329.9                 \\\hline
	\multicolumn{8}{c}{P2: Min. Compute Cost s.t. RAM Limit}                                                                                                   \\ \hline
	\multirow{5}{*}{$P_{max}$}  & 16 kB   & 15.199          & 1412.3          & 17.904          & 931.3          & \multicolumn{2}{c}{\multirow{2}{*}{(No Solution)}} \\
	& 32 kB   & \gray{25.803} & \gray{1266.3} & \gray{26.24}  & \gray{769.2} & \multicolumn{2}{l}{}                               \\
	& 64 kB   & \gray{63.603} & \gray{1121.7} & \gray{45.283} & \gray{684.6} & 63.456                   & 9458.6                  \\
	& 128 kB  & 83.133          & 947.0           & 89.6            & 683.4          & 94.224                   & 5370.3                  \\
	& 256 kB  & 181.44          & 879.2           & \multicolumn{2}{c}{(SAA)}        & 247.808                  & 3923.2                 \\ \hline
\end{tabular}
\end{table}

\subsection{Impact on Power Consumption}
We used the Nordic Semiconductor Power Profiler Kit II (PPK2) on nrf52840 boards to provide preliminary measurements of the power draw. We observed that across all optimization configurations, inference current and deep sleep current remained consistently around 5.4 mA and 1.9 mA, respectively, which hints at a straightforward link between latency and energy consumption. We note, however, that the precision of the inference current we measured may be limited due to our measurement setup and should be subject to further investigations.

\section{Network Architecture Extension}
\label{appdix:nn_archi_ext}

We are currently extending msf-CNN to support RNN- and Transformer-based architecture.

\subsection{msf-CNN on RNN/LSTM/GRU}

For 1-D sequence input, it is trivial to fuse the cascade RNN cells into one block, to avoid outputting the complete sequence between RNNs during inference. This eliminates RAM usage for storing intermediate output, especially when facing a long input sequence. The compute cost remains the same as the original models, since no recompute was introduced compared to 2-D inputs.

For 2-D input \cite{xingjian2015convolutionallstm}, all matrix-vector multiplications are replaced by 2-D convolutions and the Hadamard product is applied on two matrices inside the RNN cells. In this case, we can analyze the compute graph to check which convolution operators (inter- or intra RNN cells) can be fused into blocks, and calculate their RAM usage, compute cost as we did in the original msf-CNN.

\subsection{msf-CNN on Transformer}

We only considered decoder-only architecture so far.

For 1-D sequence input, there is not much optimization space, since the transformer/attention blocks are full of dense layers or heavy matrix-matrix multiplications, where patch-based fusion cannot be applied to reduce RAM usage. On the other hand, we could apply msf-CNN on the covolutional layers before or after transformer blocks of hybrid model (such as in ViT \cite{guo2023visual}), and integrate Efficient Attention \cite{shen2021efficient} to further improve space and compute complexity of attention layer.

For 2-D image input, msf-CNN can be extended to the convoluional variants of common attention layers (such as VAN \cite{guo2023visual} and SegNeXt \cite{guo2022segnext}) which contain cascade convolutional layers inter- or intra the attention layers. In this case, msf-CNN can be applied on deciding the fusion strategy.

\end{document}